\pdfoutput=1

\documentclass[11pt]{article}

\usepackage[]{naacl2021}

\usepackage{times}
\usepackage{latexsym}

\usepackage[T1]{fontenc}

\usepackage[utf8]{inputenc}

\usepackage{microtype}

\newcommand{\unanswerable}{\ensuremath{\epsilon}}
\newcommand{\text}{T}
\newcommand{\target}{S}
\newcommand{\source}{D}

\title{QuestEval: Summarization Asks for Fact-based Evaluation}

\author{Thomas Scialom$^{\star \ddagger}$,  Paul-Alexis Dray$^{\star}$, Patrick Gallinari$^{\ddagger}$, Sylvain Lamprier$^{\ddagger}$,
\\{\bf Benjamin Piwowarski$^{\diamond \ddagger}$, Jacopo Staiano$^{\star}$, Alex Wang$\dagger$} \\
$^\diamond$ CNRS, France\\
$^\ddagger$ Sorbonne Universit\'e, CNRS, LIP6, F-75005 Paris, France\\
$^\star$ reciTAL, Paris, France \\
$^\dagger$ New York University \\

  {\tt thomas@recital.ai }}

\usepackage{graphicx}
\usepackage{comment}
\usepackage{booktabs}
\usepackage{url}
\usepackage{enumitem}
\usepackage{float}

\usepackage{pifont}
\newcommand{\cmark}{\ding{51}}
\newcommand{\xmark}{\ding{55}}%

\begin{document}

\maketitle
\begin{abstract}

Summarization evaluation remains an open research problem: current metrics such as ROUGE are known to be limited and to correlate poorly with human judgments. To alleviate this issue, recent work has proposed evaluation metrics which rely on question answering models to assess whether a summary contains all the relevant information in its source document. 
Though promising, the proposed approaches have so far failed to correlate better than ROUGE with human judgments. 

In this paper, we extend previous approaches and propose a unified framework, named \texttt{QuestEval}.
In contrast to established metrics such as ROUGE or BERTScore, \texttt{QuestEval} does not require any ground-truth reference. Nonetheless, \texttt{QuestEval} substantially improves the correlation with human judgments over four evaluation dimensions (consistency, coherence, fluency, and relevance), as shown in the extensive experiments we report. 

\end{abstract}

\section{Introduction}

The reliability of automatic evaluation metrics is an important factor for progress in artificial intelligence tasks, enabling the comparison and improvement of proposed systems.
The design of reliable metrics for natural language generation (NLG) systems is very challenging, and still an open research problem:
\citet{novikova2017we, peyrard2019studying} showed that current metrics do not correlate well with human judgments, and argued for the development of new evaluation metrics. 

Among NLG tasks, summarization is one of the most difficult to evaluate automatically. For a given document, the number of possible correct outputs is much larger than for other tasks such as machine translation. 
Thus, when only a single reference summary is given -- as is typically the case for large-scale summarization datasets, the correlation of standard automatic evaluation metrics with human judgments is low \cite{louis2013automatically}. 
Furthermore, since a summary must be shorter than the corresponding source document, information selection \cite{li-etal-2018-improving} is critical so that the summary only contains the salient contents from its source document.
For these reasons, $n$-gram based metrics, such as ROUGE \cite{lin2004rouge}, are known to poorly reflect human preference \cite{louis2013automatically, novikova2017we, paulus2017deep, Bhandari2020Metrics}.
Finally, it is crucial for reliable summarization to generate texts that are factually consistent with their source documents. However, this aspect is not measured by $n$-grams based metrics.
Notably, while recent state-of-the-art generative models \cite{lewis2019bart, zhang2019pegasus} produce fluent summaries, they frequently contain false or unsupported information \cite{kryscinski2019evaluating}, a phenomenon also known as neural hallucination \cite{rohrbach2018object, zhao2020reducing}. 

To overcome these limitations, a new approach to evaluate summarization systems has recently emerged, based on question generation (QG) and answering (QA) \cite{chen2017semantic, scialom2019answers, eyal2019question}. These metrics measure to which extent a summary provides sufficient information to answer questions posed on its corresponding source document. They can be used to assess the factual consistency (i.e. precision) \cite{durmus2020feqa, wang2020asking} or the relevance (i.e. recall) \cite{scialom2019answers} of the evaluated summary, with respect to its source document. 
Although these works have introduced an interesting and novel method to evaluate summarization, with encouraging preliminary results, 
none of those metrics is found to perform better than ROUGE \cite{fabbri2020summeval}: automatic evaluation of summarization systems remains an open research problem \cite{kryscinski-etal-2019-neural}.

\begin{figure*}[!ht]
    \centering
    \includegraphics[width=0.9\linewidth]{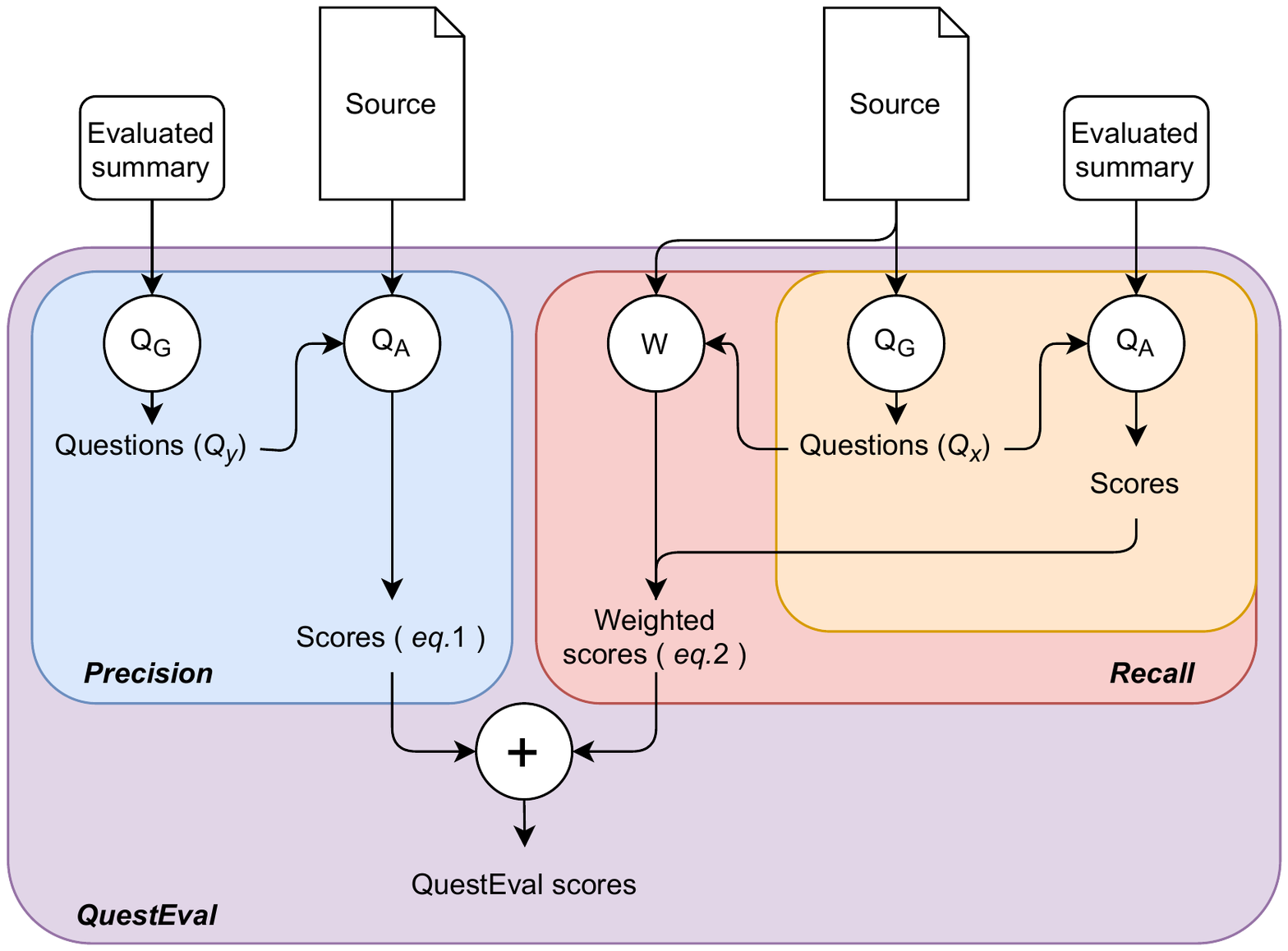}
    \caption{Illustration of the \texttt{QuestEval} framework: the blue area corresponds to the precision-oriented framework proposed by \citet{wang2020asking}. The orange area corresponds to the recall-oriented SummaQA \cite{scialom2019answers}. We extend it with a weighter component for an improved recall (red area). The encompassing area corresponds to our proposed unified approach, \texttt{QuestEval}. }
    \label{fig:safeval_schema}
\end{figure*}

Inspired by these works, and motivated to take up the challenge of summarization evaluation, we propose  \texttt{QuestEval}, a new \emph{reference-less} evaluation metric, which is found to correlate dramatically better with humans judgments. Our contributions are as follows:

\begin{itemize}
    \item We show that, by unifying the precision and recall-based QA metrics, we obtain a more robust metric;
    \item We propose a method to learn the saliency of the generated queries,  allowing to integrate the notion of information selection; 
    \item We evaluate \texttt{QuestEval} on two corpora containing annotated summaries from CNN/Daily Mail \cite{nallapati2016abstractive} and XSUM \cite{narayan2018don} datasets. The proposed metric obtains state-of-the-art results in terms of correlation with humans judgments, over all the evaluated dimensions. 
    Notably, \texttt{QuestEval} is effective at measuring factual consistency, a crucial yet challenging aspect for summarization.

\end{itemize}

\section{Related Work}

\paragraph{Summarization Metrics}
The most popular evaluation metric for summarization is ROUGE \citep{lin2004rouge}, which computes the recall of reference $n$-grams in the evaluated summary. 
Other $n$-grams based metrics have been proposed such as CIDEr \cite{vedantam2015cider} and METEOR \cite{lavie-agarwal-2007-meteor}, but none of them correlates better with humans according to SummEval, a recent large study conducted by \citet{fabbri2020summeval}.

Recent works have leveraged the success of pretrained language models.
\citet{zhang2019bertscore} proposed BERTScore, which uses BERT \citep{devlin2018bert} to compute a similarity score between the reference and the evaluated text. 
However, its performance is similar to that of ROUGE \citep{fabbri2020summeval}. 
Several works have explored using natural language inference (NLI) models to evaluate the factual consistency of summaries \citep{kryscinski2019evaluating,falke2019ranking,maynez2020faithfulness}, finding mixed results in using NLI models rather than QA models.

\paragraph{QA-Based Metrics}

QA-based approaches for summary evaluation were proposed a decade ago by \citet{clarke2010discourse} for human evaluation. \citet{chen2017semantic} and \citet{eyal2019question} proposed to automate this approach by automatically generating questions from the reference summary. 
\citet{scialom2019answers} extended these works by generating the questions from the source document, which probes information recalled from the input text in the output summary, and thus is recall oriented. However, by weighing each question equally, their approach lacks a way to select questions that reflect the most important information of the input.
Conversely, \citet{wang2020asking} and \citet{durmus2020feqa} proposed to generate questions from the evaluated summary. These methods are precision oriented, since they measure the amount of information in the evaluated summary that are supported by the input text. 
We show in this paper that combining these recall and precision approaches leads to an improved metric.

\section{A Question-Answering based Framework} 
\label{QAQG}

This paper introduces the \texttt{QuestEval} framework for evaluating summarization systems, that accounts for both factual consistency and relevance of the generated text, without requiring any human reference.
\texttt{QuestEval} consists of a QG component $Q_G$ and a QA component $Q_A$, described in this section and depicted in Figure~\ref{fig:safeval_schema}. 

\subsection{Question Answering}
\label{sec:QA}

Recently, there has been significant progress on factoid question answering, with models obtaining human-level performance on benchmarks such as SQuAD \cite{rajpurkar2016squad}. 
Leveraging on these advancements, our $Q_A$ component consists of a pretrained T5 model \cite{raffel2019exploring}, which extracts answers from a source document given the document and a question to answer. In the following, we refer to $Q_A(r|\text,q)$ as the probability of the answer $r$ to question $q$ on a text $\text$, and $Q_A(\text,q)$ as the answer greedily generated from the model.     

When a summary is evaluated, there is no guarantee that it contains the answer. Therefore, it is crucial for the QA model to be able to predict when a question is unanswerable. Our $Q_A$ component thus includes the \emph{unanswerable} token, that we denote \unanswerable, among its possible outputs.

\subsection{Question Generation}
\label{subsec:QG}
For the QG component, we draw on recent work on neural answer-conditional question generation \cite{zhou2017neural}.
The component also consists of a T5 model, finetuned to maximize the likelihood of human questions, given the corresponding answer and source document.

At test time, given a source document or generated summary, we first select a set of answers from the text to condition the QG model on.
Following \citet{wang2020asking}, we consider all the named entities and nouns from the source document as answers. 
Then, for each selected answer, we generate a question via beam search.\footnote{We experimented with nucleus sampling \cite{holtzman2019curious} to increase diversity of the questions, with no success.}
We filter out every question for which the QA model
predicts an incorrect answer.
Based on this, we denote $Q_G(\text)$ the set of question-answer pairs $(q,r)$ for a text $T$ such that $Q_A(\text, q) = r$.

\section{The QuestEval metric}

In the following, $\source$ and $\target$ are two sequences of tokens with $\source$ denoting the source document and $\target$ the corresponding evaluated summary. 

\subsection{Precision}

A summary is deemed inconsistent with respect to its source text if, given a question, the answer differs when conditioned on $\target$ or $\source$.
Therefore, we define the precision score for the evaluated summary as: 
\begin{equation}\small
    \label{eq:precision}
    Prec(\source, \target) = \frac{1}{|Q_G(\target)|} \sum\limits_{(q,r) \in Q_G(\target)} F1(Q_A(\source,q), r) 
\end{equation}
The F1 score is a standard metric for evaluating factoid question answering models, and measures the overlap between the predicted answer and the corresponding ground truth. 
It outputs $1$ for an exact match between both answers and $0$ if there is no common token. 
This definition of factual consistency corresponds to the frameworks concurrently proposed by \citet{wang2020asking} and \citet{durmus2020feqa}.

\subsection{Recall}
\label{subsec:Recall}
While a summary should contain only factual information (precision), it should also contain the most important information from its source text (recall). Extending \citet{scialom2019answers} by introducing a query weighter $W$, we define recall as:
\begin{equation}\small
    \label{eq:recall}
    Rec(\source, \target) =
    \frac{
        \sum\limits_{q,r \in Q_G(\source)}
        W(q, \source) 
        (1-Q_A(\unanswerable|\target, q)) 
    }{
        \sum\limits_{q,r \in Q_G(\source)} W(q, \source)
    }
\end{equation}

\noindent where $Q_G(\source)$ is the set of all question-answer pairs for the source text $\source$,  
and $W(q,\source)$ is the weight of query $q$ for text $\source$.

\paragraph{Answerability and F1} 

Factoid question answering models are commonly evaluated using F1 score, measuring the overlap between the predicted answer and the corresponding ground truth \cite{rajpurkar2016squad}. However, an answer could be correctly expressed in different ways, e.g. ``ACL'' and ``Association for Computational Linguistics''. Unfortunately, the F1 score is $0$ in this example. 

To sidestep this issue, \citet{scialom2019answers} use the QA confidence of answerability, i.e. $1 - Q_A(\epsilon)$, rather than F1 score. 
Defining recall this way allows to measure  answerability independently of the way the answer is expressed, 
but does not take into account possible model hallucinations, i.e. the summary could answer the question incorrectly.

Conversely, when we assess factual consistency, it is not enough for a question from the summary to be answerable from the source document. The two answers to this question should \emph{also share the same meaning} to be factually consistent. 
While using answerability allows for more true positive (e.g. ``ACL''), for precision it is crucial to detect true negatives.
This motivates our use of the F1 score in this case, similar to \citet{wang2020asking}.

\paragraph{Query Weighting}

In \citet{scialom2019answers}, all questions are considered equally important,
i.e. the weight $W(q,\source) = 1$ for every query
$q \in Q_G(\source)$. 
However, since a summary necessarily has a constrained length, an effective summary should contain the most important information from the source.
To account for this, we introduce a question weighter, which is trained to distinguish \emph{important} questions from \emph{anecdotal} ones. 
We leverage existing summarization datasets to create training data for the weighter:
given a source document $\source$, each question $q \in Q_G(\source)$ is labeled as \emph{important} if the corresponding human summary contains the answer, as computed by the QA component applied on the summary (i.e. $Q_A(\target, q)\neq\unanswerable$).

$W(q,\source)$ denotes the probability that $q$ is important for $\source$. 
Note that the question weighter only concerns recall, and therefore is not applied when computing precision. 

\subsection{Unifying Precision and Recall}
The final \texttt{QuestEval} score accounts for both the precision and recall by computing their harmonic mean (i.e. the F-Score): $2\frac{Prec . Rec}{Prec + Rec}$. The \texttt{QuestEval} score is thus directly comparable with existing evaluation metrics, such as ROUGE or BLEU, as it lies in the same numerical range.

\section{Experiments}

\begin{table*}[hbt!]
\centering
\begin{tabular}{lrrrrrrr}
				
                          & \#Ref & Consistency & Coherence & Fluency        & Relevance      & Average          \\
                          
ROUGE-1                   & 11 & 18.1          & 20.1          & 14.9          & 35.6          & 22.2 \\
ROUGE-L                   & 11 & 15.7          & 15.6          & 13.8          & 33.4          & 19.6 \\
METEOR                    & 11 & 3.3           & 2.9             & 7.1           & -0.5          & 3.2  \\
BLEU                      & 11 & 17.5           & 22.           & 13.7          & 35.6          & 22.2 \\
BERTScore-f               & 11 & 20.3           & 18.5          & 21.6           & 31.9          & 23.1  \\
\midrule
ROUGE-1                   & 1 & 11.0 & 9.8 & 7.5 & 18.9 & 11.8 \\
ROUGE-L                   & 1 & 8.2 & 7.3 & 5.7 & 13.5 & 8.7 \\
BLEU                      & 1 & 8.9 & 3.9 & 4.0 & 12.7 & 7.4 \\
BERTScore-f               & 1 & 8.7 & 9.8 & 10.6 & 17.9 & 11.8 \\
\midrule
SummaQA                   & 0 & 8.3            & 8.0           & -2.9          & 26.2          & 9.9  \\
QAGS (our impl.)          & 0 & 20.4           & 7.7           & 16.8          & 9.1           & 13.7 \\
\midrule
\texttt{QuestEval}$_{W=uniform}$                  & 0 & 43.7            & 22.9           & 28.2           & 37.5           & 33.1  \\
\hfill {\scriptsize \textit{w/o QA neg sampl.}} & 0 & 42.5            & 22.5           & 27.7           & 37.2           &  32.4          \\
\texttt{QuestEval}$_{W=learned}$                  & 0 & 42.0            & \textbf{24.0}  & 28.4           & \textbf{39.2}  & \textbf{33.5}  \\
\hfill {\scriptsize \textit{Precision Only}}    & 0 & \textbf{46.5}   & 14.0           & \textbf{30.9}  & 22.2           & 28.4  \\
\hfill {\scriptsize \textit{Recall Only}}       & 0 & 30.5            & 22.6           & 19.2           & 37.6           & 27.5  \\
\bottomrule

\end{tabular}
\caption{Summary-level Pearson correlation coefficients for various dimensions between automatic metrics and human judgments on SummEval. The top section corresponds to correlations for metrics computed on 11 reference summaries, as reported in \citet{fabbri2020summeval}. The second section corresponds to these metrics, but given only one reference. The third section corresponds to the QA-based baselines. The bottom section corresponds to the proposed \textit{QuestEval} and its ablations.}
\label{tab:main_res_cnn}
\end{table*}

\subsection{Summarization Datasets}
\label{sec:dataset}

To evaluate \texttt{QuestEval}, we measure its correlation with human judgments on different datasets:

\paragraph{SummEval} Released by \citet{fabbri2020summeval}, it is one of the largest human-annotated datasets for summarization. Derived from CNN/Daily Mail \cite{nallapati2016abstractive}, it consists of 12.800 summary level annotations. To ensure diversity, the summaries were generated from 16 different summarization models, including extractive and abstractive architectures. To ensure quality, three experts annotated four dimensions:
\emph{i) Consistency}: the proportion of facts in the summary corresponding to facts in the original text;
\emph{ii) Coherence}: how well-structured and well-organized is the summary; 
\emph{iii) Fluency}: how fluent the summary is to read;
and, \emph{iv) Relevance}: the ratio between important and excess information in the summary.\footnote{
See 4.3 Human Annotations in \citet{fabbri2020summeval} for more details.}

\paragraph{QAGS-XSUM} \citet{wang2020asking} made available a subset of 239 BART outputs \cite{lewis2019bart} fine-tuned on XSUM \cite{narayan2018don}.\footnote{Note that XSUM provides more abstractive summaries than those of CNN/Daily Mail.} Three annotators measured the ``correctness'' of each summary, which corresponds to consistency in SummEval.

\subsection{Question Answering \& Generation} 
\label{subsec:QA_and_QG}
To train our $Q_G$ and $Q_A$ models, we used two factoid question answering datasets: SQuAD-v2 \cite{rajpurkar2018know} and NewsQA \cite{trischler2016newsqa}. Such datasets are composed of (paragraph, question, answer) triplets. SQuAD-v2 provides unanswerable questions, while NewsQA is composed of news articles, corresponding to our summarization domain. 
Note that QG can be seen as the dual task for QA. Any QA dataset can be reversed into a QG dataset, by switching the generation target from the answer to the question.

Lastly, we found it helpful to train our QA model using additional synthetic unanswerable questions. 
This is done by considering a shuffled version of  the dataset, where each question is randomly assigned to a paragraph from another triplet of the dataset. We consider these additional samples, with flipped contexts, as unanswerable. 
All experiments, except otherwise specified, use this additional negative sampling process to improve identification of unanswerable queries.   

\subsection{Baselines Metrics}

As baselines, we considered the following: 

\paragraph{N-gram based}
ROUGE \cite{lin2004rouge} is the most widely used evaluation in summarization. This metric measures the recall of reference n-grams in the evaluated summary. 
Conversely, BLEU \cite{papineni2002bleu} computes the precision of summary n-grams in the references. METEOR \cite{lavie-agarwal-2007-meteor} is a variant that uses stemming, synonyms and paraphrastic matches.

\paragraph{Neural based}
Leveraging recent progress in language modeling, \citet{zhang2019bertscore} proposed BERTScore: for each token of the summary, the maximum cosine similarity is computed over contextualized token embeddings of the reference summary, and the overall mean is reported.

\paragraph{Question based}
SummaQA \cite{scialom2019answers} is a recall oriented metric, with questions generated from the source document. QAGS \cite{wang2020asking} is a precision oriented metric, with questions generated from the summary. 

\subsection{Results}

In Tables~\ref{tab:main_res_cnn} and~\ref{tab:main_res_xsum} we report the results for \texttt{QuestEval}, along with several ablations.
$W=uniform$ corresponds to setting all questions weights equal.
Conversely, $W=learned$ corresponds to the weights learned as detailed in Section~\ref{subsec:Recall}. We also report the recall and precision component separately.

In Table~\ref{tab:main_res_cnn}, we observe that, amongst existing metrics, BERTScore achieves the best average Pearson correlation with human judgements (23.1), slightly above ROUGE-1 (22.2) and BLEU (22.2). 
These correlations are obtained when providing \emph{no less than 11 gold references}, and averaging results. Given  a single reference, all these correlations are halved. Most of the large scale datasets provide only one reference per example in their test set (e.g. CNN/Daily Mail and XSUM), a fact that highlights the importance of searching for more reference-efficient alternatives.

With regards to sample efficiency, QA-based metrics \emph{do not require any references}. We expect Relevance to be better measured by Recall oriented metrics, and less so for Consistency. This is confirmed in the results, where SummaQA correlates better with Relevance than Consistency (26.2 vs 8.3), and vice versa for QAGS (9.1 vs 20.4). 
By unifying and extending the two, \texttt{QuestEval} allows to take both dimensions into account, improving the average correlation by 18\% (28.4 to 33.5).

The dimension that benefits the most from the learned question weighter is Relevance (+4\%, from 37.5 to 39.2), indicating that our classifier learns \emph{which questions target important information}. We discuss this aspect more in depth in the following section.

Finally, compared to the other metrics, the improvement is remarkable (33.5 vs 11.8 for BERTScore), and allows safer evaluations of the systems while not even requiring references. 

\begin{table}
\centering
\begin{tabular}{lr}
Metric          & Consistency \\
\midrule
ROUGE-1         & 13.2                                                   \\
ROUGE-L         & 8.9                                                    \\
METEOR          & 10.0                                                   \\
BLEU            & 5.6                                                    \\
BERTScore       & 2.5                                                    \\
\midrule
SummaQA        & -                                                       \\
QAGS     & 17.5                                                   \\
\midrule
\texttt{QuestEval}$_{W=uniform}$                  &  30.4  \\
\hfill {\scriptsize \textit{w/o QA neg sampl.}} &  28.5   \\
\texttt{QuestEval}$_{W=learned}$                  &  29.0   \\
\hfill {\scriptsize \textit{Precision Only}}    &  \textbf{32.7}  \\ 
\hfill {\scriptsize \textit{Recall Only}}       &  13.9  \\
\bottomrule
\end{tabular}

\caption{Summary-level Pearson correlation coefficients for Correctness between various automatic metrics and human judgments on QAGS-XSUM. The top section corresponds to correlations for diverse metrics computed on one reference summary, as reported in \citet{wang2020asking}. The middle section corresponds to QA-based baselines. The bottom section corresponds to this work.}
\label{tab:main_res_xsum}
\end{table}

\subsection{Discussion}

\paragraph{Reference-less}

\begin{figure*}
    \centering
    \includegraphics[width=.45\linewidth]{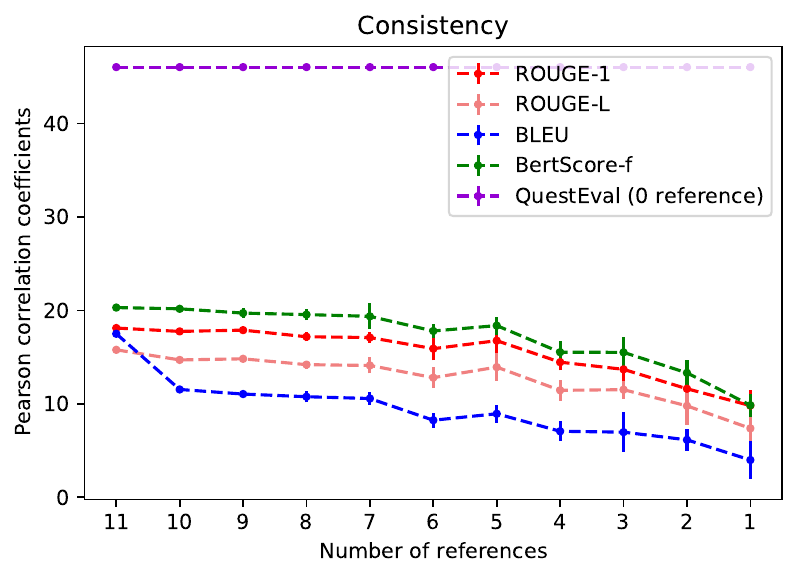}
    \includegraphics[width=.45\linewidth]{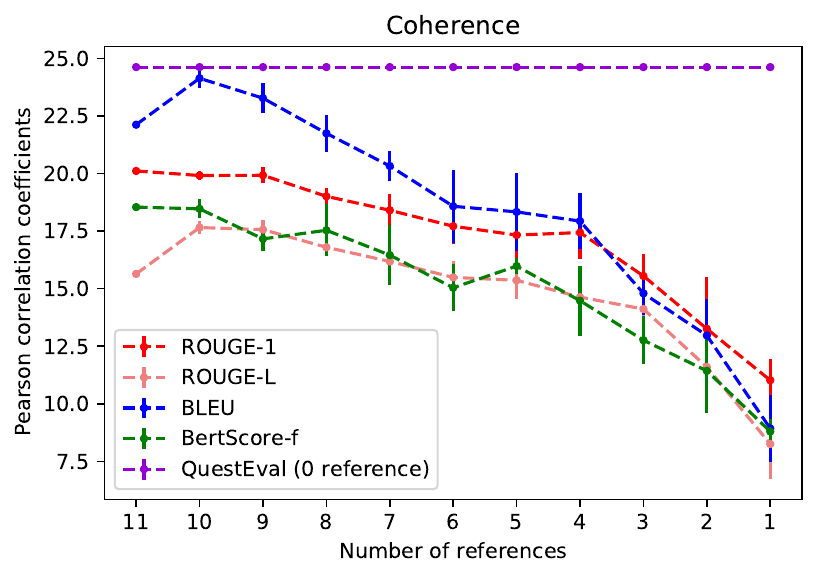}
    \includegraphics[width=.45\linewidth]{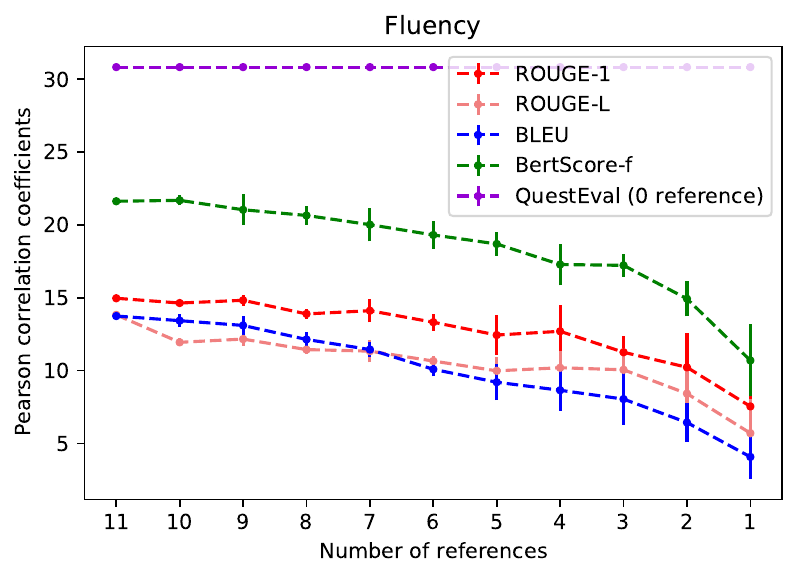}
    \includegraphics[width=.45\linewidth]{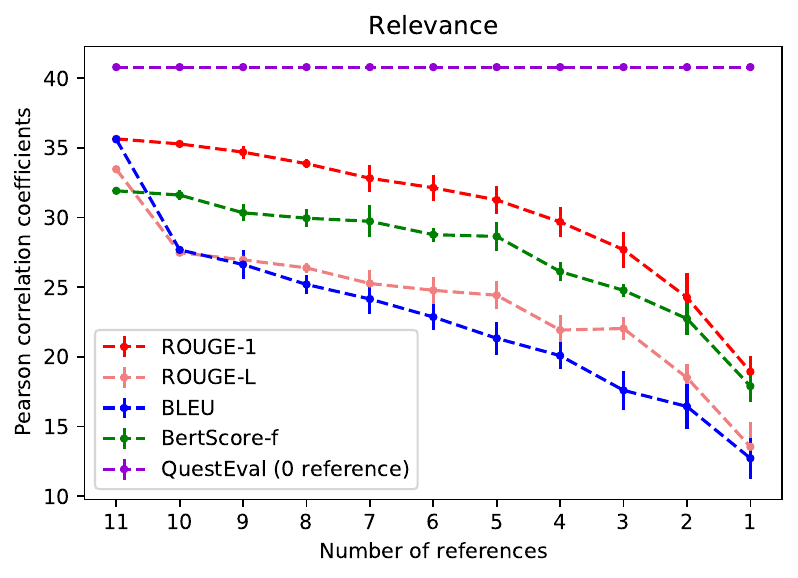}
    \caption{Variation of the Pearson correlations between various metrics and humans, versus the number of references available. \texttt{QuestEval} is constant, since it is independent from the references.}
    \label{fig:pearson_corels_wrt_N}

\end{figure*}

One of the main limitations for the current metrics is that they require gold references to compute similarity scores. However, many possible summaries are valid for one source document. We argue that the universe of correct outputs is much larger than in other generation tasks such as machine translation. This explains why the correlations with humans is largely reduced when computed with one reference instead of 11 (see Table~\ref{tab:main_res_cnn}: BERTScore-f drops from 23.1 to 11.8 in average, and other metrics likewise). 
Unfortunately, assuming the availability of as many as 11 gold references is not realistic in most scenarios, due to the cost of obtaining reference summaries.  

To complement Table~\ref{tab:main_res_cnn},
we report in Figure~\ref{fig:pearson_corels_wrt_N} the correlations for the best baselines as we progressively decrease the number of available gold references from 11 to 1. We observe that for all four dimensions and all the baselines, the correlations decrease and the variance increases as the number of references decreases. 
However, \texttt{QuestEval} \emph{does not require any reference}. Therefore, the improvement over the other metrics grows larger as the number of references used decreases. Furthermore, \texttt{QuestEval} enables the evaluation of systems on datasets even if \emph{no gold reference is available}.

\paragraph{Query Weighter}
\phantomsection
\label{para:question_generation_weighter}

\begin{table}[]
\begin{center}
\begin{tabular}{ccr}
                  \emph{important}  & \emph{answered}                     & Relevance Corr. \\
\midrule
\cmark                    &  \cmark                   & 37.6 \\
\cmark                    &  \xmark                   & -33.5 \\
\xmark                    &  \cmark                   & -5.7  \\
\bottomrule
\end{tabular}
\caption{Pearson correlation coefficients between human judgments (for Relevance) and the percentage of \emph{important} and/or \emph{answered} questions, on SummEval data.}
\label{tab:set_question_w_correls}
\end{center}

\end{table}

There is no unique answer to the question ``What makes a good summary?'': it depends on the reader's point of view, which makes summarization evaluation challenging. For instance, given a contract, the seller and the buyer could be interested in different information within the same document.

In this paper, to instantiate the weighter $W$, we propose to learn a specific dataset policy: ``what kind of questions are likely answered in the CNN/Daily Mail training summaries?" This is a reasonable heuristic given that editors created the summaries following their specific policy.

To demonstrate the effectiveness of the weighter, we proceed as follows. We first consider that a question $q\in Q_G(\source)$, generated on the source document, is \textit{important} if the probability given by the query weighter is above a threshold, i.e. if $W(\source, q)>0.5$. 
We then say that a question is \textit{answered} if the probability of being unanswerable is below a threshold, i.e. $Q_A(\unanswerable | \target, q) <0.5$.
Therefore, a question can belong to one of four folds, given the two above criteria (\textit{important} and/or \textit{answered}).
In Table~\ref{tab:set_question_w_correls}, we measure how the percentage of questions belonging to a specific fold correlates with the Relevance dimension for each generated summary on SummEval. 
We observe that the percentage of questions that are \textit{important} and \textit{answered} correlates positively with Relevance, as opposed to the percentage of questions that are \textit{important} but not \textit{answered}. 
Finally, the percentage of questions that are \textit{answered} but not \textit{important} does not correlate with Relevance. It indicates that our proposed approach is able to learn what are the questions that should be asked or not.

We emphasize that $W$ is a flexible component of our framework. It can be adapted to specific domains and applications. 
For instance, one could design a specific $W$, to focus the evaluation on information about specific entities, such as people or events. 

\paragraph{An Explainable Metric}

\begin{table}
    \centering\small
    \begin{tabular}{p{.94\linewidth}}
    \toprule
    \textbf{Source Document} This is the embarrassing moment a \textit{Buckingham Palace} guard slipped and fell on a manhole cover in front of hundreds of shocked tourists as he took up position in his sentry box. 
    [...]
    The Guard comprises two detachments, one each for Buckingham Palace and St James's Palace, under the command of the Captain of The Queen's Guard.  \\
    
    \textbf{Generated Question} Where was the Changing of the Guard held? \\
    
    \textbf{Weighter prediction} \textit{Important Question} \\

    \textbf{Answer Span} \textcolor{blue}{Buckingham Palace} \\

    \midrule
    \textbf{Correct Summary} The Queen's Guard slipped on a manhole cover during the Changing of the Guard at \textit{Buckingham Palace} last week. [...] \\

    \textbf{Predicted Answer} \textcolor{green}{Buckingham Palace}: \cmark \\

    \midrule
    \textbf{Hallucinated Summary} The Queen's Guard slipped on a manhole cover during the Changing of the Guard at \textit{St James's Palace} last week. [...]
    
    \textbf{Predicted Answer} \textcolor{red}{St James's Palace}: \xmark \\
    
    \midrule
    \textbf{Incomplete Summary} The Queen's Guard slipped on a manhole cover during the Changing of the Guard during an embarrassing moment.. [...]

    \textbf{Predicted Answer} \textcolor{red}{Unanswerable}: \xmark \\
    
    \bottomrule
    \end{tabular}
    \caption{Sample output from \texttt{QuestEval}: a generated question, it's predicted importance given a source document; the corresponding predicted answers to the question, for three different summaries.}
    
    \label{tab:example_safeval}
\end{table}

One important feature of \texttt{QuestEval} is its explainability. It is straightforward to investigate \emph{1)} what are the important points not answered in the summary and \emph{2)} what are the inconsistencies between the source document and the summary.
We illustrate this in Table~\ref{tab:example_safeval}, with a source document, from which a question $q$ is generated and answered. According to the weighter $W$, $q$ is categorized as \textit{important}. Three evaluated summaries are then shown.

The first summary $\target\textsubscript{correct}$ is factually consistent with the source document: the predicted answer $Q_A( \target\textsubscript{correct}, q)$ corresponds to the source document answer \emph{Buckingham Palace}. 
The second summary $\target\textsubscript{hallu}$ is factually inconsistent with the source document: the predicted answer $Q_A(\target\textsubscript{hallu}, q)$ \emph{does not} correspond to \emph{Buckingham Palace}.
Finally, the third summary $\target\textsubscript{incomplete}$  does not answer the question, i.e. $Q_A(\target\textsubscript{incomplete}, q)=\unanswerable$, and
is thus incomplete.

\paragraph{Negative Sampling Effect}

\begin{figure}[!t]
    \centering
    \includegraphics[width=\linewidth]{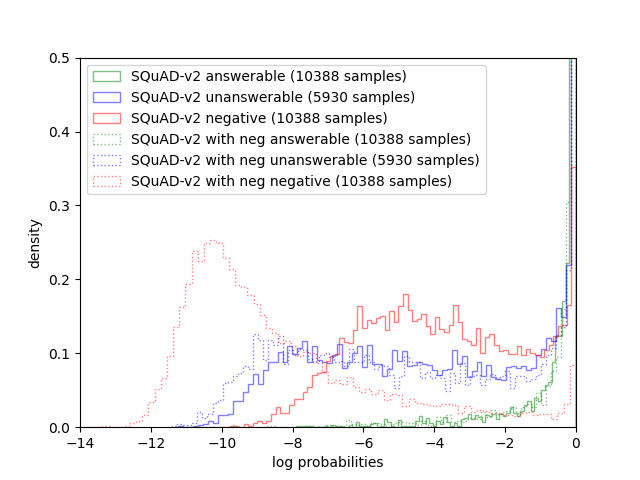}
    \caption{Distribution of the log probabilities of answerability -- i.e. $\log (1- Q_A(\unanswerable|T, q))$ -- for two QA models. 1) solid lines: a model trained on SQuAD-v2 \emph{without} the negative sampled examples. 2) dashed lines: a model trained on SQuAD-v2 \emph{with} the negative sampled examples. The evaluated samples belong to three distinct categories: 1)  answerable, 2) unanswerable questions (but present in SQuAD-v2) and 3) the negatively sampled ones (as described in Section~\ref{sec:dataset}).}
    \label{fig:distribution_qa_neg}
\end{figure}

In Tables~\ref{tab:main_res_cnn} and~\ref{tab:main_res_xsum}, when \texttt{QuestEval} uses a QA model trained without negative sampling (see Section~\ref{subsec:QA_and_QG}), we observe a decrease of performance, from 33.3 to 32.4 on SummEval and from 30.4 to 28.5 on QAGS-XSUM. 

In Figure~\ref{fig:distribution_qa_neg}, we report the distribution of the log probabilities for the two QA models, trained with and without negative sampling. We can observe that the QA model exposed to the negative sampling during training, has learned to separate better the negative sampled questions (for negative, i.e. red lines, the dashed line is more on the left than the solid line).

Indeed, the unanswerable questions of SQuAD-v2 were written adversarially by crowd-workers, to look similar to answerable ones. However, in the context of \texttt{QuestEval}, unanswerable questions are not adversarial. It simply often happens that the summary does not contain the answer.
Therefore, \texttt{QuestEval} sees unanswerable questions that look like the one we built trough our negative sampling method, rather than the adversarial ones. 
This may explain the improvement of a \texttt{QuestEval} with a QA model trained with negative sampling.

\paragraph{Computational Complexity}

\begin{figure}[!ht]
    \centering
    \includegraphics[width=\linewidth]{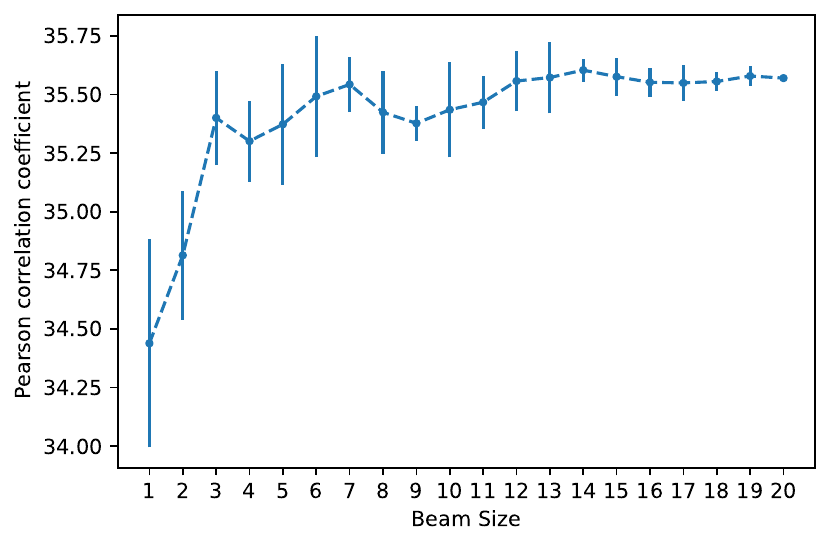}
    \caption{Pearson correlation with humans on SummEval w.r.t. the QG beam size. 
    }
    \label{fig:complexity}
\end{figure}

Following \citet{wang2020asking}, we generate the questions with $K=20$ beams during decoding and we keep all the different versions of the questions in the latter steps, which improves correlations. However, the downside of this is the inference time which increases linearly w.r.t the beam size. 
To be widely adopted, a metric should not only correlate with human judgment, but also be computationally efficient. 
In Figure~\ref{fig:complexity} we show the variation of the average correlation with respect to the beam size. 
The improvement from $K=1$ to $K=20$ is small (34.4 to 35.6), and the rank order for the different systems remains unchanged. 
Therefore, we believe that using \texttt{QuestEval} with $K=1$ is a reasonable choice, allowing for fast computation while preserving the correlation with human judgments.

\section{Conclusion}

We proposed \texttt{QuestEval}, a new \emph{reference-less} framework to evaluate summarization models, which unifies and extends previous QA-based approaches
with question weighting and negative sampling,
accounting for 
factual consistency, relevance and information selection.

We implement \texttt{QuestEval} leveraging state-of-the-art deep learning models.
Compared to existing metrics, we find that \texttt{QuestEval} correlates dramatically better with human judgments, while at the same time not requiring any gold reference. This allows for more accurate comparison between systems. 
Moreover, any progress in question answering and generation can directly be applied to our proposed approach, leading to further potential improvements.
We make the code available\footnote{\url{https://github.com/recitalAI/QuestEval}} with the hope that it will contribute to further progress in the field.

We are currently adapting QuestEval to other Natural Language Generation tasks that suffer from the same evaluation limitations, such as machine translation and text simplification. 
In future work, we plan to extend \texttt{QuestEval} to a multilingual version.

\section*{Acknowledgments}
This work was partially performed using HPC resources from GENCI-IDRIS (Grant 2021-AD011011841).

\bibliography{naaclhlt2019}
\bibliographystyle{acl_natbib}

\appendix

\end{document}